\documentclass[11pt,journal,onecolumn]{IEEEtran}
\usepackage{amsmath,amssymb,amsfonts,latexsym,amsthm,enumerate,fullpage,hyperref,dsfont,subcaption}
\usepackage{stackrel,color,mathtools}
\usepackage[noadjust]{cite}
\mathtoolsset{showonlyrefs=true}
\newtheorem{theorem}{Theorem}

\newtheorem{definition}{Definition}
\newtheorem{corollary}{Corollary}

\def\Expt{\mathds{E}}

\def\argmax{\mathop{\rm argmax}}
\newcommand{\dfn}{\triangleq}

\newcommand{\indfunc}[1]{\mathds{1}_{\left(#1\right)}}
\newcommand{\Acond}{p^{\mathrm{cond}}}
\newcommand{\qAcond}{q^{\mathrm{cond}}}
\newcommand{\Bern}{\text{Ber}}

\begin{document}
\title{{Constructing Multiclass Classifiers using Binary Classifiers Under Log-Loss}}

\author{Assaf Ben-Yishai and Or Ordentlich 
	\thanks{ 
	The authors are with the School of Computer Science and Engineering, Hebrew University of Jerusalem, Israel.
	Emails: \{assafbster@gmail.com,or.ordentlich@mail.huji.ac.il\}
	This work was supported by an ISF grant no. 1791/17.
	This paper was presented in part at ISIT 2021.}}
\maketitle

\begin{abstract}
The construction of multiclass classifiers from binary elements is studied in this paper, and performance is quantified by the regret, defined with respect to the Bayes optimal log-loss. We discuss two known methods. The first is one vs. all (OVA), for which we prove that the multiclass regret is upper bounded by the sum of binary regrets of the constituent classifiers. The second is hierarchical classification, based on a binary tree. For this method we prove that the multiclass regret is exactly a weighted sum of constituent binary regrets where the weighing is determined by the tree structure.

We also introduce a leverage-hierarchical classification method, which potentially yields smaller log-loss and regret. The advantages of these classification methods are demonstrated by simulation on both synthetic and real-life datasets.
\end{abstract}

\section{Introduction}
Classification is one of the most basic problems in statistical learning. The  classification task is the following: given some data (e.g. an image) provide a guess to the class from which it was taken (e.g. the object appearing in the image). The guess can be the label of the chosen class, which will be referred to as \textit{hard} classification. The guess can also be a probability attached to each one of the class labels, which will be referred to as \textit{soft} classification.

Clearly, the minimal non-trivial number of classes is two. For this reason, there is abundant research and practice for the binary classification problem, including many readily available implementations. 
On the other hand, many classification problems are by nature multiclass, (i.e. having more than two classes) and their solution requires the construction of multiclass classifiers.
For some classifier families (such as logistic regression), the multiclass counterpart can be constructed by a straightforward generalization of the simple binary case. For other families however, (such as support vector machines) there is no single straightforward generalization. {For such cases, it is desirable to have systematic methods for reducing the construction of a multiclass classifier to that of constructing several binary classifiers and then fusing them {into a single  classifier}
\cite{daniely2012multiclass,rifkin2004defense,hastie1998classification,dietterich1994solving,allwein2000reducing,kumar2002hierarchical,chen2004integrating,vural2004hierarchical,lorena2008review,delmoral2021pitfalls}. 

Since black-box implementations of various binary classification architectures are so prevalent, such a modular solution is highly {appealing}.
{In addition}, building a multiclass classifier from binary elements can be beneficial even if a simple multiclass generalization exists. The motivation in these cases can be to provide a binary based architecture which better fits the data structure or reduces the learning and recognition complexity \cite{morin2005hierarchical,silla2011survey}.}

The main difference between the current contribution and the existing literature is the use of {the logarithmic loss (log-loss) as the target loss function}. {Namely, the focus in the works cited above is on the construction of multiclass classifiers from binary elements and performance analysis under hard classifiers and the zero-one loss (i.e.~classification error probability).}
Instead, we consider soft classifiers and quantify their performance by assessing the quality of their probability assignment using log-loss \cite{merhav1998universal,cesabianchilugosi,cw14,jcvw15,fogel2018universal,xr20}. More precisely, we evaluate the \emph{regret}, which is the excess log-loss compared to the optimally attained one. The use of soft classifiers and log-loss is particularly suitable in the high error probability regime. In other words - for difficult classification problems in which even an expert cannot give an accurate guess and therefore only assigns posterior probabilities to the classes.

Two methods for constructing {multiclass classifiers from binary elements} are discusses and analyzed. The first method is the well know one-vs.-all (OVA). In this method there is a binary classifier for every class, discriminating it from a set of all the remaining ones. The final probability assignment is the normalization of the probabilities of these binary classifiers. We show that the total regret of the OVA is upper bounded by the sum of the regret of the constituent binary classifiers (Theorem~\ref{theorem:OVAregret} and Corollary~\ref{cor:OVAcond}). 

The second method is called hierarchical classification. It is based on a binary tree architecture, with a binary classifier assigned at each node. For this method, we show that the total regret is identical to a weighted sum of the regrets of the constituent binary classifiers and the weighing is determined by the tree structure (Theorem~\ref{theorem:treeregret} and Corollary~\ref{cor:treecond}). 

{Our analysis of the regret corresponding to the known OVA and hierarchical classification methods, while fairly simple, reveals important insights on the relations between binary and multiclass classification. In particular, for hierarchical classification our analysis illustrates that once 
{it is decided} to use the logarithmic loss as the figure of merit, the problem of multiclass classification with $K$ possible classes is completely \emph{equivalent} to $K-1$ binary classification problems. This new insight motivates us to develop a novel architecture termed \textit{leveraged} hierarchical classification. In this architecture, which is a variant of hierarchical classification, one starts with a multiclass classifier (e.g., Softmax), and improves it. This is attained by allowing the binary classifiers at the tree nodes of a hierarchical classifier to have the same parametric richness as that of the original multiclass classifier, but without coupling the parameters at different nodes. We show that this method, if judiciously trained, can yield an improvement  over directly using a single instance of the multiclass classifier in terms of the total log-loss.}

We conclude the paper with experimental results which use Softmax based classifiers. We use two datasets. The first is a synthetic dataset in which the actual log-loss and regret are calculated, and the generalization error is evaluated. We show that the leveraged-hierarchical architecture improves upon the simple multiclass classifier in terms of log-loss. The second dataset is the {popular} MNIST dataset \cite{deng2012mnist} for handwritten digits. In this dataset we show that a leveraged hierarchical classifier yields a substantial improvement in terms of both error rate and log-loss with respect to the Softmax benchmark.

The rest of the paper is organized as follows: Section~\ref{sec:formul} give the problem formulation. Section~\ref{sec:OVA} defines the OVA method and states our related results. Section~\ref{sec:gentree} defines the hierarchical classification method and states our related results. Section~\ref{sec:training} explains how the constituent binary classifiers are learned using a training set, and Section~\ref{sec:leveraging} introduces our novel leveraged classifier architecture and its training. Finally, Section~\ref{sec:numeric} give the experimental result, and Section~\ref{eq:conclusion} gives some concluding remarks.

\section{Problem Formulation \label{sec:formul}}
The statistical setting of the standard classification problem is characterized by the pair of dependent random variables: $X$ and $Y$ which are drawn from a (possibly unknown) distribution $P_{XY}$. The random variable $Y\in\mathcal{Y}=\{0,...,K-1\}$ corresponds to class label and $X\in\mathcal{X}$ corresponds to the observation. The goal is to come up with a classifier $f(X)$ that is close to $Y$ with respect to some loss function. The most common loss function is the zero-one loss, and the corresponding classifier is designed such as to minimize the classification error probability $\Pr(f(X)\neq Y)$. In many cases, however, the observation $X$ reveals some information on the label $Y$, but not enough to accurately predict the label. In such cases, a preferable approach is to design classifiers that output soft information, namely a conditional probability distribution for $Y$ given $X$, rather than committing to a single value of $Y$. The common choice for a loss function measuring the quality of such a ``soft classifier'' is the logarithmic loss (log-loss). 

Let us now define the log-loss and its associated regret. For simplicity of exposition, we start by introducing the notation without conditioning on the observation $X$. The generalization to the conditional case is given in the sequel. As defined above, $Y\sim P$ is the class random variable supported on  $\mathcal{Y}=\{0,\ldots,K-1\}$, and $K>2$ (i.e.~multiclass).  
We use $P(i)$ and $p_i$ interchangeably to denote the class probability 
$\Pr(Y=i)$. Let $Q$ denote a (possibly mismatched) probability assignment on $\mathcal{Y}$.
We define the log-loss for predicting $Y$ based on $Q$, while the actual underlying distribution is $P$ by
\begin{align}\label{eq:defL}
    L({P},Q) \dfn  \Expt_{y\sim P}\log\frac{1}{Q(y)}.
\end{align}
{where ``$\log$" denotes the natural logarithm}.
This quantity is also known as the \emph{cross-entropy} of $Q$ relative to $P$.
{It is well-known that $\min_Q L(P,Q)$ is attained by $Q=P$, and that for this choice $L(P,P)=H(Y)$, where $H(\cdot)$ denotes entropy function.
We therefore} denote the {regret} related to using $Q$ instead of $P$ by
\begin{align}\label{eq:defR}
    R(P,Q) \dfn  L(P,Q)-L(P,P)=D(P\mid\mid Q),
\end{align}
where $D(P\mid\mid Q)$ is the Kullback-Leibler divergence between $P$ and $Q$. 
We shall use $R(P,Q)$ and $D(P\mid\mid Q)$ interchangeably, where the first notation shall be used to state results, and the latter shall be used for the analysis. 

In the sequel, binary classifiers will associated with Bernoulli random variables. For clarity of notation we use lowercase letters to denote the properties of these Bernoulli random variables, such as success probability, log-loss, regret etc. Namely, let $U\sim\Bern(p)$, 
(i.e. $U\in\{0,1\}$, and $\Pr(U=1)=p$), and let $q$ be the parameter of a possibly mismatched distribution $\Bern(q)$. 
The related (binary) log-loss is denoted by
\begin{align}
    \ell({p},q) \dfn  p\log\frac{1}{q}+(1-p)\log\frac{1}{1-q},
\end{align}
and the related binary regret is denoted by
\begin{align}
    r(p,q) \dfn  \ell(p,q)-\ell(p,{p})= d(p\mid\mid q),
\end{align}
where 
\begin{align}
d(p\mid\mid q)\dfn p\log\frac{p}{q}+(1-p)\log\frac{1-p}{1-q}    
\end{align}
denotes the binary divergence. $\ell({p},p)$ is equal to the binary entropy of $p$ denoted by
\begin{align}
h(p)\dfn p\log \frac{1}{p}+(1-p)\log\frac{1}{1-p}.
\end{align}

Let us now extend the notation to the standard classification case, in which we are interested in predicting $Y$ according to some observation $X$. The pair $(X,Y)$ is distributed according to
\begin{align}
(X,Y)\sim P_{XY}=P_X\times P_{Y|X}.
\end{align}
The observation random variable $X$ is supported on $\mathcal{X}$, where $\mathcal{X}$ is either some discrete alphabet or $\mathcal{X}=\mathbb{R}^d$. We denote the posterior probability of the class $y$ given the observation $x$ by
\begin{align}
P_{Y\mid X=x}(y)  \dfn \Pr(Y=y\mid X=x).
\end{align}
We denote the estimated posterior probability of the class $y$ given the observation $x$ provided by a soft classifier by $Q_{Y|X=x}$. The expected conditional log-loss of $Q_{Y|X}$ is defined as
\begin{align}\label{eq:condL}
    L(P_{Y\mid X},Q_{Y\mid X}\mid P_X) \dfn
    \Expt_{x\sim P_{X}}
    L(P_{Y\mid X=x},Q_{Y\mid X=x})
\end{align}
and the conditional regret is defined similarly
\begin{align}\label{eq:condR}
    R(P_{Y\mid X},Q_{Y\mid X}\mid P_X) &\dfn
    \Expt_{x\sim P_{X}}
    R(P_{Y\mid X=x},Q_{Y\mid X=x}).
\end{align}
Note that {$R(P_{Y\mid X},Q_{Y\mid X}\mid P_X)=D(P_{Y\mid X}\mid\mid Q_{Y\mid X}\mid P_X)$, 
where $D(P_{Y\mid X}\mid\mid Q_{Y\mid X}\mid P_X)=\Expt_{x\sim P_{X}} D(P_{Y\mid X=x}\mid\mid Q_{Y\mid X=x})$ is the conditional divergence.}
The Bernoulli counterparts are appropriately defined as follows
\begin{align}
    \ell({p}_{\mid X},q_{\mid X}\mid P_X) &\dfn  \Expt_{x\sim P_{X}}
    \ell({p}_{\mid X=x},q_{\mid X=x})\\
    r({p}_{\mid X},q_{\mid X}\mid P_X) &\dfn  \Expt_{x\sim P_{X}}
    r({p}_{\mid X=x},q_{\mid X=x})
\end{align}


\section{One vs.~all (OVA)\label{sec:OVA}}
For the sake of simplicity of exposition, we start with the unconditional
case, and add the conditioning in the sequel.
We standardly use $\indfunc{\cdot}$ to denote an indicator function, being equal to one if the condition is satisfied and zero otherwise. Using indicator functions we can define the following set of Bernoulli random variables related to $Y$
\begin{align}
A_i=A_i(Y)\dfn \indfunc{Y=i},~~~i=0,\ldots,K-1.
\end{align}
Trivially, 
\begin{align}
p_{A_i}\dfn\Pr(A_i=1)=P(i).
\end{align}
The identity implies that the set of success probabilities $\{p_{A_i}\}_{i=0}^{K-1}$  uniquely describe the distribution of $Y$. The one vs.~all (OVA) method uses a set of $K$ binary classifiers (related to the set of indicators $\{A_i\}_{i=0}^{K-1}$), each discriminating a specific class from all the rest. It can be compactly defined as follows.
\begin{definition}[One vs. all (OVA)] \label{def:ova}
Given a set of $K$ estimates $\{q_{A_i}\}_{i=0}^{K-1}$, {not all zero}, of the respective probabilities $\{\Pr(A_i=1)\}_{i=0}^{K-1}$,
the OVA estimate of $P$ is defined as
\begin{align}\label{eq:OVAdef}
Q^{\mathrm{OVA}}(i)=\frac{q_{A_i}}{\sum_{j=0}^{K-1}q_{A_j}},~~~i=0,\ldots,K-1.
\end{align}
\end{definition}

Let us now state our main results for the OVA method which relates its regret to the regrets of its constituent binary classifiers $\{r(p_i,q_{A_i})\}$.
\begin{theorem}[OVA regret]\label{theorem:OVAregret}
\begin{align}\label{eq:RegretIneq}
{R}({P},Q^{\mathrm{OVA}})\leq \sum_{i=0}^{K-1}r({p}_{i},q_{A_i})
\end{align}
\end{theorem}
\begin{proof}
We start by rewriting \eqref{eq:OVAdef} as $Q^{\mathrm{OVA}}(i)=\frac{q_{A_i}}{\alpha K}$, for $i=0,\ldots,K-1$,
where $\alpha \dfn \frac{\sum_{i=0}^{K-1}q_{A_i}}{K}$. Note that since $q_{A_i}\in[0,1]$ for all $i$ it is guaranteed that $\alpha\in(0,1]$ (note
that by Def.~\eqref{def:ova}, $\{q_{A_i}\}$ are not all zero, so
$\alpha>0$).
Recalling that the regrets can be written as divergences, the statement in \eqref{eq:RegretIneq} is equivalent to
\begin{align}
\sum_{i=0}^{K-1}d({p}_{i}\mid\mid q_{A_i})-D({P}\mid\mid Q^{\mathrm{OVA}})\geq 0.
\end{align}

Expanding $D({P}\mid\mid Q^{\mathrm{OVA}})$ yields
\begin{align}\label{eq:RPY}
D({P}\mid\mid Q^{\mathrm{OVA}}) &= \sum_{i=0}^{K-1}p_i\log\frac{p_i}{q_{A_i}/(\alpha K)}\\
&=\sum_{i=0}^{K-1}p_i\log\frac{p_i}{q_{A_i}}+\log(\alpha K),
\end{align}
and expanding $\sum_{i=0}^{K-1}d({p}_{i}\mid\mid q_{A_i})$ yields
\begin{align}
\sum_{i=0}^{K-1}&d({p}_{i}\mid\mid q_{A_i}) \\
=&\sum_{i=0}^{K-1}\left(p_i\log\frac{p_i}{{q_{A_i}}}
+(1-p_i)\log\frac{1-p_i}{1-{q_{A_i}}}
\right).\label{eq:RPYsum}
\end{align}
Subtracting \eqref{eq:RPY} from \eqref{eq:RPYsum} we obtain
\begin{align}\label{eq:Rsum}
\sum_{i=0}^{K-1}d({p}_{i}\mid\mid& q_{A_i})-D({P}\mid\mid Q^{\mathrm{COVA}})\\
=& \sum_{i=0}^{K-1}(1-p_i)\log\frac{1-p_i}{1-{q_{A_i}}}-\log (\alpha K)\\
=& F_1 +F_2,
\end{align}
where the last transition is by adding and subtracting the term  
$(K-1)\log\frac{K-1}{K(1-\alpha)}$ and defining 
\begin{align}
F_1 \dfn
\sum_{i=0}^{K-1}(1-p_i)\log\frac{1-p_i}{1-{q_{A_i}}}
-(K-1)\log\frac{K-1}{K(1-\alpha)}
\end{align}
and 
\begin{align}
F_2\dfn(K-1)\log\frac{K-1}{K(1-\alpha)}-\log (\alpha K).
\end{align}

We start by proving that $F_1\geq 0$ by using the log-sum inequality for the sets  
$\{1-p_i\}_{i=0}^{K-1}$ and $\{1-q_{A_i}\}_{i=0}^{K-1}$. We note that 
$p_i\leq 1$ and ${q_{A_i}}\leq 1$ so it is guaranteed that the sets comprise only non-negative elements. For the singular case of $p_i=1$ we use the convention that $0\log 0=0$. Applying the log-sum inequality \cite[Theorem 17.1.2]{CoverThomas}, as explained above, yields
\begin{align}
&\sum_{i=0}^{K-1}(1-p_i)\log\frac{1-p_i}{1-{q_{A_i}}}\\
&- \left(\sum_{i=0}^{K-1}(1-p_i)\right)\log\frac{\sum_{i=0}^{K-1}(1-p_i)}
{\sum_{i=0}^{K-1}(1-q_{A_i})}
\geq 0\label{eq:lsi1}.
\end{align}
Note that $\sum_{i=0}^{K-1}p_i=1$ and $\sum_{i=0}^{K-1}{q_{A_i}}=\alpha K$, which implies that $\sum_{i=0}^{K-1}(1-p_i)=K-1$ and $\sum_{i=0}^{K-1}(1-q_{A_1})=K(1-\alpha)$. Plugging the values of the sums in \eqref{eq:lsi1} yields
\begin{align}
&\sum_{i=0}^{K-1}(1-p_i)\log\frac{1-p_i}{1-{q_{A_i}}}\\
&- \left(K-1\right)\log\frac{K-1}{K(1-\alpha)}
\geq 0
\end{align}
which proves that $F_1\geq 0$ by its definition.\newline
Let us now rearrange $F_2$
\begin{align}
    F_2&=
(K-1)\log\frac{K-1}{K(1-\alpha)}-\log (\alpha K)\\
&=K\left(
\left(1-\tfrac{1}{K}\right)\log\frac{1-\tfrac{1}{K}}{(1-\alpha)}
+\tfrac{1}{K}\log \frac{\tfrac{1}{K}}{\alpha}\right)\\
&=K\cdot d\left(\tfrac{1}{K}\mid \mid \alpha\right)\geq 0.
\end{align}
The last inequality stems from the non-negativity of the divergence, implies that $F_2\geq 0$ hence concludes the proof of the theorem.
\end{proof}

The following corollary extends Theorem~\ref{theorem:OVAregret} to the conditional case
\begin{corollary}[OVA conditional regret]\label{cor:OVAcond}
\begin{align}
{R}({P_{Y\mid X}},Q_{Y\mid X}^{\mathrm{OVA}}\mid P_X)\leq \sum_{i=0}^{K-1}r({p}_{i\mid X},q_{A_{i}\mid X}\mid P_X).
\end{align}
\end{corollary}
\begin{proof}
We start by using Theorem~\ref{theorem:OVAregret} where all the probabilities are under the  point-wise conditioning $X=x$
\begin{align}
{R}({P_{Y\mid X=x}},Q_{Y\mid X=x}^{\mathrm{OVA}})\leq \sum_{i=0}^{K-1}r({p}_{i\mid X=x},q_{A_{i}\mid X=x}).
\end{align}
Taking the expectation  w.r.t $x\sim P_X$ on both sides yields
\begin{align}
&\Expt_{x\sim P_X}\left({R}({P_{Y\mid X=x}},Q_{Y\mid X=x}^{\mathrm{OVA}})\right)\\
&\leq \Expt_{x\sim P_X}\left(\sum_{i=0}^{K-1}r({p}_{i\mid X=x},q_{A_{i}\mid X=x})\right).
\end{align}
The Corollary now follows the definition of the conditional regret.
\end{proof}

Although this paper deals with prediction under logarithmic loss, it is worth mentioning that our results also have implication on prediction under zero-one loss (error probability), as stated in the following corollary
\begin{corollary}[OVA regret under zero-one loss]
Let $P_{e,\mathrm{Bayes}}$ denote the error probability attained by the optimal (maximum a-posteriori) classifier $\hat{y}_{\mathrm{MAP}}(x)=\argmax_y P_{Y|X=x}(y)$ for $y$ from $x$. Denote by $P_e(Q_{Y\mid X}^{\mathrm{OVA}})$ the error probability attained by the estimator $\hat{y}_{\mathrm{OVA}}(x)=\argmax_y Q^{\mathrm{OVA}}_{Y|X=x}(y)$. Then,
\begin{align}
P_e&(Q_{Y\mid X}^{\mathrm{OVA}})-P_{e,\mathrm{Bayes}}\\&\leq \sqrt{2\sum_{i=0}^{K-1}r({p}_{i\mid X},q_{A_{i}\mid X}\mid P_X)}.
\end{align}
\end{corollary}
\begin{proof}
Let $\mathrm{TV}(P,Q)$ denote the total variation distance between the distributions $P$ and $Q$. The mismatched error probability for predicting $Y\sim P$ based on maximum a-posteriori with respect to the incorrect distribution $Q$ satisfies
\begin{align}
&P_e(Q)=1-P(\argmax_{y}Q(y))\\
&=P_e(P)+P(\argmax_{y}P(y))-P(\argmax_{y}Q(y))\label{eq:tvmis}\\
&\leq P_e(P)+2\mathrm{TV}(P,Q),
\end{align}
where the last transition is by adding the term $Q(\argmax_{y}Q(y))-Q(\argmax_{y}P(y))\geq 0$ and using the definition of $\mathrm{TV}(P,Q)$. We can therefore write
\begin{align}
P_e(Q^{\mathrm{OVA}}_{Y\mid X})&-P_{e,\mathrm{Bayes}}\leq2\Expt_{x\sim P_{X}}\mathrm{TV}\left(P_{Y|X=x},Q^{\mathrm{OVA}}_{Y\mid X=x} \right)\\
&\leq2\Expt_{x\sim P_{X}} \sqrt{\frac{1}{2} D\left(P_{Y\mid X=x}\mid\mid Q^{\mathrm{OVA}}_{Y\mid X=x}\right)} \\
&\leq \sqrt{2 R\left(P_{Y\mid X},Q^{\mathrm{OVA}}_{Y\mid X}\mid P_X\right)},
\end{align}
where we have used Pinsker's inequality \cite[Lemma 17.3.2]{CoverThomas} in the second inequality  and Jensen's inequality in the third. The statement now immediately follows by bounding $R\left(P_{Y\mid X},Q^{\mathrm{OVA}}_{Y\mid X}\mid P_X\right)$ according to Corollary~\ref{cor:OVAcond}.
\end{proof}

\section{Hierarchical Classification \label{sec:gentree}}
\begin{figure}[ht]
\centering
	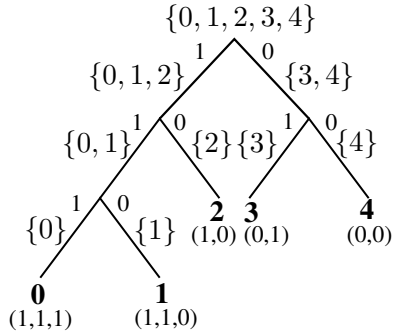
	\caption{A tree for five classes\label{fig:noncovatree}}
\end{figure}

\begin{figure}[ht]
	\centering
	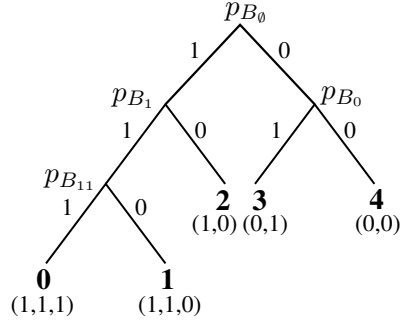
	\caption{The tree of Fig.~\ref{fig:noncovatree} with Bernoulli probabilities\label{fig:treeBer}}
\end{figure}
The second method discussed in this contribution is hierarchical classification. It is based on constructing the multiclass classifier 
from binary classifiers which are assigned to the internal (i.e. not leaf) nodes of a binary tree. Each tree node corresponds to a partition of its set of classes into two disjoint sets. Every bifurcation at the tree nodes is also associated with a bit, which gives rise to an equivalent and useful representation of a prefix code. It is also easy to see that every nested binary partition of $K$ classes can be represented using exactly $K-1$ splits. Therefore, it is sufficient to consider trees having exactly $K-1$ internal nodes.

For example, consider the tree with five classes depicted in Fig.~\ref{fig:noncovatree}. Every internal node has a set of classes which is split into two disjoint subsets with respect to a bit value. For example, the root node splits the set $\{0,1,2,3,4\}$ into the subsets $\{0,1,2\}$ and $\{3,4\}$ corresponding to a bit values $1$ and $0$ respectively. In this way, each class is associated with a codeword (i.e. a bit sequence). Namely, classes $0-4$ are respectively associated with  codewords $(1,1,1)$, $(1,1,0)$, $(1,0)$, $(0,1)$ and $(0,0)$. 

In order to calculate the class probabilities, every tree node is associated with a Bernoulli success probability, which represents the probability of the node bit to be equal to one. The success probability of the node corresponding to the prefix $c$ (i.e. the string of bits generated along the path from the tree root to the node) is denoted by $p_{B_c}$. The root itself is associated with the empty prefix $\emptyset$ and its Bernoulli probability is denoted by $p_{B_\emptyset}$. 
An example to this concept is given in 
Fig.~\ref{fig:treeBer} which denotes the node Bernoulli probabilities of the tree in Fig.~\ref{fig:noncovatree}. The one-to-one mapping between the class probabilities and the binary node probabilities is given in the following examples
\begin{align}
    p_{B_\emptyset} = \Pr(Y\in\{0,1,2\}\mid Y\in\{0,1,2,3,4\}),
\end{align}
\begin{align}
    p_{B_1} = \Pr(Y\in\{0,1\}\mid Y\in\{0,1,2\}),
\end{align}
\begin{align}
P(1)=p_{B_{\emptyset}}\cdot p_{B_1}\cdot (1-p_{B_{11}}),
\end{align}
and
\begin{align}
P(4)=(1-p_{B_{\emptyset}})\cdot (1-p_{B_{0}}).
\end{align}

In addition, we also use a short hand indexing of the tree nodes, referring to every node using an integer in the set $\{0,...,K-2\}$, without indicating its prefix.
Then, the set of all descendant classes of node $j$ is denoted by $S_j$. The set of descendant classes respective to the branch labeled $0$ (resp.~labeled $1$) is denoted by $S_j^0$ (resp.~$S_j^1=S_j\setminus S_j^0 $).
The success probability related to node $j$ is denoted by $p_{S_j}$, being equal to
\begin{align}
    p_{S_j}\dfn \Pr(Y\in S_j^1\mid Y\in S_j).
\end{align}

\begin{figure*}[ht]
\begin{subfigure}{0.5\textwidth}
\hspace{-1cm}
    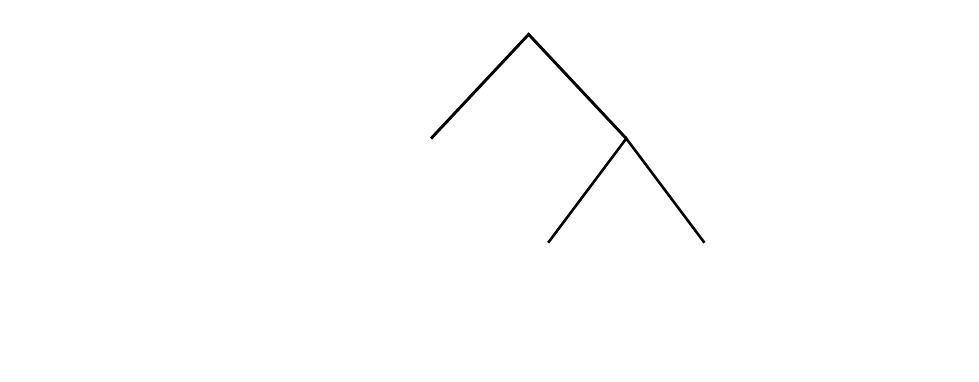
    \subcaption{True probabilities: $P$,$\{p_0,...,p_3\}$ \label{fig:treeestleft}}
\end{subfigure}
\begin{subfigure}{0.5\textwidth}
\hspace{-1cm}
     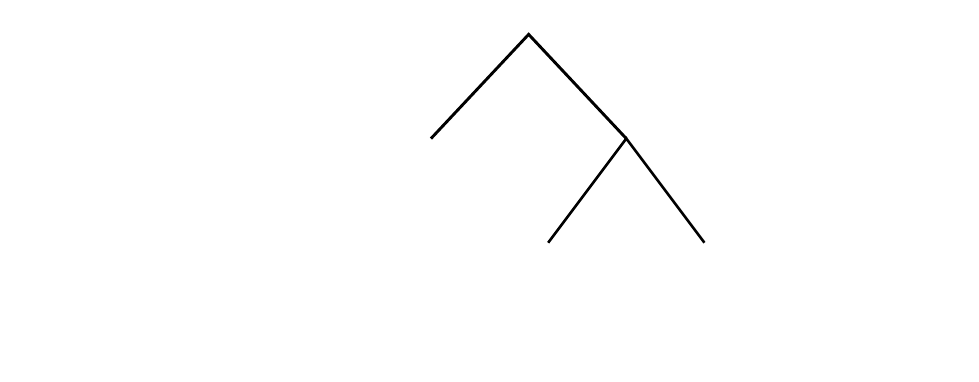	
    \subcaption{Estimated probabilities: $Q$,$\{q_0,...,q_3\}$\label{fig:treeestright}}     
\end{subfigure}
\caption{An example to a hierarchical tree with true and estimated probabilities \label{fig:treeest}}
\end{figure*}
An example to a possible indexing of the tree in Fig.~\ref{fig:treeestleft} is given in Fig.~\ref{fig:treeBer}. The class probability $P$ is built according to the tree structure and the set $\{p_{S_i}\}_{i=0}^3$ (for example $P_1 = p_{S_2}\cdot p_{S_1}\cdot(1-p_{S_0})$). \newline

We are interested in devising an estimate $Q$ for the class probability assignment $P$ using a fixed tree structure and a set of estimated binary success probabilities  $\{q_{S_i}\}_{i=0}^{K-2}$, one for each node.
For example, consider the tree in Fig.~\ref{fig:treeestright} which represents an estimated version of the tree in Fig.~\ref{fig:treeestleft}.
For instance, in this tree, $Q(1)$ is an estimate of $P(1)$ and is calculated using the estimates to the Bernoulli probabilities $q_{S_0},q_{S_1}$ and $q_{S_2}$ according to $Q(1) = q_{S_2}\cdot q_{S_1}\cdot(1-q_{S_0})$ whereas the true probability is 
$P(1) = p_{S_2}\cdot p_{S_1}\cdot(1-p_{S_0})$.
The following theorem relates the regrets associated with the binary probability estimates to the regret of the resulting multiclass probability assignment.
\begin{theorem}[Hierarchical tree regret]\label{theorem:treeregret}

{For a fixed tree structure, for which $P$ induces the node success probabilities $p_{S_0},\ldots,p_{S_{K-2}}$, and an hierarchical classifier using the same tree with node success probabilities $q_{S_0},\ldots,q_{S_{K-2}}$, the obtained distribution $Q$ satisfies}
\begin{align}\label{eq:RPQ}
{R}({P}, Q)=\sum_{j=0}^{K-2}\Pr(Y\in S_j)r(p_{S_j},q_{S_j}).
\end{align}
\end{theorem}
\begin{proof}
\begin{figure}[ht]
	\centering
	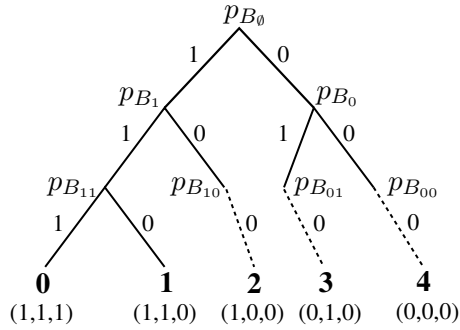
 	\caption{The tree of Fig.~\ref{fig:treeBer} after zero padding.\label{fig:noncovazero}}
\end{figure}

We denote the tree depth, i.e. the length of the longest codeword, by $\mathcal{D}$, where $\lceil\log_2 K\rceil\leq \mathcal{D}\leq K-1$. 
We then extend all the codewords to length $\mathcal{D}$ by appropriate zero padding and adding dummy nodes with a zero Bernoulli success probability. This process is illustrated in Fig.~\ref{fig:noncovazero}. 
We denote the probability distributions $P$ and $Q$ induce on the codebook $B_0^{\mathcal{D}-1}$ by $P_{B_0^{\mathcal{D}-1}}$ and its appropriate estimate $Q_{B_0^{\mathcal{D}-1}}$. 
Using this notation we can write the following series of equalities
\begin{align}
&R(P,Q)=D(P\mid\mid Q)\\
&=D(P_{{B_0^{\mathcal{D}-1}}}\mid\mid Q_{{B_0^{\mathcal{D}-1}}})\label{eq:D1}\\
&=\sum_{i=0}^{\mathcal{D}-1}
D\left(P_{B_i\mid{B}_0^{i-1}}\mid\mid Q_{B_i\mid{B}_0^{i-1}}
\mid P_{{B}_0^{i-1}}\right) \label{eq:D2}\\
&=\sum_{i=0}^{\mathcal{D}-1}
\sum_{{b}_0^{i-1}\in\{0,1\}^i}\Pr({B}_0^{i-1}={b}_0^{i-1})\\
&\quad\quad\cdot D\left(P_{B_i\mid{B}_0^{i-1}={b}_0^{i-1}}\mid\mid Q_{B_i\mid{B}_0^{i-1}={b}_0^{i-1}}\right)\\
&=\sum_{i=0}^{\mathcal{D}-1}
\sum_{{b}_0^{i-1}\in\{0,1\}^i}\Pr({B}_0^{i-1}={b}_0^{i-1})
\\
&\quad\quad\cdot d\left(p_{B_{{b}_0^{i-1}}}\mid\mid q_{B_{{b}_0^{i-1}}}\right)\label{eq:prelast}\\
&=\sum_{j=0}^{K-2}\Pr(Y\in S_j)r(p_{S_j},q_{S_j}).\label{eq:Dlast}
\end{align}
where~\eqref{eq:D1} is due to the one-to-one mapping between the class labels and the codewords in the codebook, \eqref{eq:D2} is by the chain rule of the divergence, 
\eqref{eq:prelast} is by the notation of the Bernoulli probabilities on the internal nodes. Lastly \eqref{eq:Dlast} is by the alternative indexing of the internal nodes and by tree structure. Note
that all the dummy nodes (outside the set of $K-1$ indexed nodes) are associated with a zero Bernoulli probability in both $p$ and $q$. Namely, for their respective codewords $a$, we have $p_{B_a}=q_{B_a}=0$. Therefore, the contribution of these nodes to the total regret is zero.
\end{proof}
It is interesting to note that somewhat similar results appear in the literature for prefix codes (for example \cite[Equation (1)]{yeung1991local}). The main difference between our setting and the prefix code setting is that in prefix codes the Bernoulli probabilities in the simulating tree are {all set to $1/2$, whereas in the classification setup they can take arbitrary values in $[0,1]$.}

{We further remark that the generally loose, but simple upper bound on the regret
\begin{align}
{R}({P}, Q)\leq \sum_{j=0}^{K-2}r(p_{S_j},q_{S_j}),
\end{align}
can be easily derived using the data processing inequality for divergences. To see this, note that drawing $Y$ according to $P$ (resp.~ $Q$) for a specific binary tree structure can be done by drawing all the $K-1$ bits of the nodes and then
producing the prefixes by traversing the tree. The generation of the prefixes is the additional processing which reduces the divergences. 
}

The following corollary generalizes Theorem~\ref{theorem:treeregret} to the conditional case.
\begin{corollary}[Hierarchical classifier conditional regret]\label{cor:treecond}
{For a fixed tree structure, for which $P_{Y|X}$ induces the conditional node success probabilities $p_{S_0|X},\ldots,p_{S_{K-2}|X}$, and an hierarchical classifier using the same tree with node success probabilities $q_{S_0|X},\ldots,q_{S_{K-2}|X}$, the obtained distribution $Q_{Y|X}$ satisfies}
\begin{align}
&{R}({P}_{Y\mid X}, Q_{Y\mid X}\mid P_X)=\\
&\sum_{i=0}^{K-2}\Pr(Y\in S_i)r({p_{S_i\mid X}},{q_{S_i\mid X}}\mid P_{X\mid Y\in S_i}).
\label{eq:Ltreecond}
\end{align}
\end{corollary}
\begin{proof}
We begin by taking \eqref{cor:treecond}, conditioning all the probabilities by $X=x$ and taking the expectation w.r.t.~$P_X$ according to the definition
\begin{align}
&{R}({P}_{Y\mid X}, Q_{Y\mid X}\mid P_X)\\
&=\Expt_{x\sim P_X}{R}({P}_{Y\mid X=x}, Q_{Y\mid X=x})\\
&=\Expt_{x\sim P_X}\left[\sum_{i=0}^{K-2}\Pr(Y\in S_i\mid X=x)r({p_{S_i\mid X=x}},{q_{S_i\mid X=x}})\right]\\
&=\sum_{i=0}^{K-2}
\Expt_{x\sim P_X}\left[\Pr(Y\in S_i\mid X=x)r({p_{S_i\mid X=x}},{q_{S_i\mid X=x}})\right]\label{eq:summands}
\end{align}
Let us now define Bernoulli random variables $B_i\dfn \indfunc{Y\in S_i}$ and the function $g_i(x)\dfn r(p_{S_i\mid X=x},q_{S_i\mid X=x})$ and expand the $i$'th summands of \eqref{eq:summands} as follows
\begin{align}
&\Expt_{x\sim P_{X}}\left[\Pr(B_i\mid X=x)g_i(x)\right]\\
&=\Expt_{x\sim P_{X}}\left[\left[\Expt_{b\sim P_{B_i\mid X=x}}\indfunc{b}\right]g_i(x)\right]\\
&=\left[\Expt_{b\sim P_{B_i}}\indfunc{b}\right]\left[\Expt_{x\sim P_{X\mid B_i=b}}g_i(x)\right]\\
&=\Pr(B_i=1)\Expt_{x\sim P_{X\mid B_i=1}}g_i(x).\label{eq:lawoftotal}
\end{align}
Plugging \eqref{eq:lawoftotal} in back into \eqref{eq:summands}
concludes the proof. 
\end{proof}

\subsection{Conditional OVA (COVA)\label{subsec:cova}}
\begin{figure}[ht]
\centering
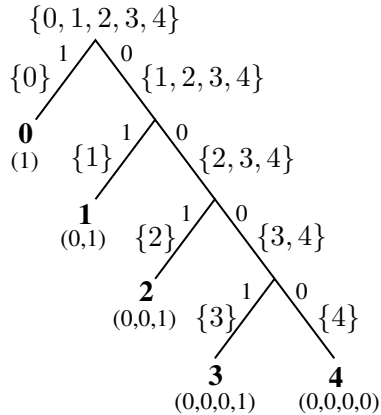
\caption{A COVA tree for five classes\label{fig:covatree}}
\end{figure}
An interesting special case of hierarchical classification which is closely related to the OVA method is the termed \textit{conditional} OVA (COVA) \cite{MulticlassISIT}. This method is inspired by the non-binary information-distilling qunatizer proposed in~\cite{bnop18} and the non-binary channel upgrading algorithm in~\cite{ot2020}.  The COVA tree has depth $K-2$ and a single binary classifier at every level. The classifier at level $i$ separates class $i$ from the set of classes $\{i+1,...,K-1\}$. Alternatively, using the hierarchical notation the COVA tree is characterized by setting
$S_i = \{i,...,K-1\}$ and $S_i^1 = i$ for all $i\in\{0,...,K-2\}$. 

An example for COVA tree with $K=5$ is depicted in Fig.~\ref{fig:covatree}. 
The COVA method can be regarded a conditioned modification of OVA in the following sense: like OVA, every binary classifier separates only a single class from a subset of other classes, but the preceding classes are omitted from the competing set. In addition, it uses only $K-1$ binary classifiers (and not $K$ as in the OVA method) and the tree structure ensures that the COVA estimate produces a valid probability assignment without requiring normalization. 

For the COVA regret statement it is useful to give the following equivalent notation fo the Bernoulli probabilities for $i\in\{0,...,K-2\}$
\begin{align}\label{eq:pAidef}
\Acond_{A_i}&\dfn\Pr(A_i=1\mid Y\geq i)
\end{align}
where $A_i=\indfunc{Y=i}$.
Denoting the estimate for $\Acond_{A_i}$ by $\qAcond_{A_i}$ and applying Theorem~\ref{theorem:treeregret} yields the following corollary
\begin{corollary}[COVA regret]\label{corr:COVAregret}
\begin{align}
    {R}({P}, Q^{\mathrm{COVA}})=\sum_{i=0}^{K-2}\Pr(Y\geq i)r(\Acond_{A_i},\qAcond_{A_i}).\label{eq:COVAregret}
\end{align}
\end{corollary}

\section{Training the Binary Classifiers \label{sec:training}}
In supervised learning under log-loss, the learner is given a training set of labeled samples $T\dfn\{(x_i,y_i)\}_{n=1}^N$ drawn independently from an unknown distribution $P_{XY}$, and is required to output a conditional distribution $Q_{Y|X}$ for which the regret $R(P_{Y|X},Q_{Y|X}{\mid P_X})$ is small. We are interested in a ``black-box'' reduction from the multiclass supervised learning problem to the binary case. To this end, assume we have access to an ``off-the-shelf" binary classifier, (e.g., logistic regression, decision tree) which takes a training set with binary labels $\{(x_n,a_n)\}_{n=1}^N$, $x\in\mathcal{X}$, $a_n\in\{0,1\}$, as input, and constructs a probability assignment $q_{A\mid X=x}$ for every $x\in \mathcal{X}$ as output. Clearly, in the general case the training set does not contain all the values in $\mathcal{X}$ (which is particularly impossible in the case that $\mathcal{X}=\mathbb{R}^d$). Nevertheless, the task of the training algorithm is to provide an estimate to the posterior probability denoted by $q_{A\mid X=x}$ to all values in $\mathcal{X}$, including the ones that are not present in $T$. For example, for logistic regression with two classes the probability assignment is given by 
\begin{align}
    q_{A\mid X=x}(0) &=\frac{\exp(\beta_0^Tx)}{\exp(\beta_0^Tx)+\exp(\beta_1^Tx)}\\
    q_{A\mid X=x}(1) &=\frac{\exp(\beta_1^Tx)}{\exp(\beta_0^Tx)+\exp(\beta_1^Tx)},
\end{align}
where $\beta_i\in\mathbb{R}^d$ are learned according to the training set $T$. We note that a common convention is to set $\beta_1$ as the all zeros vector, which has no effect the expressive power of the model. 

\begin{figure}[ht]
\centering
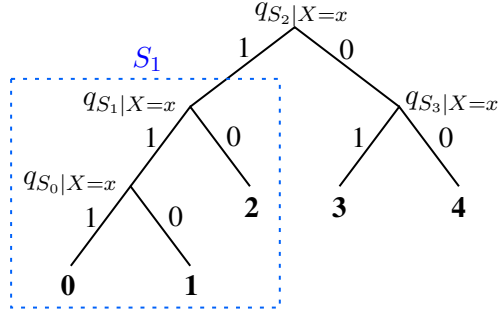
\caption{Training the binary classifier $q_{S_1\mid X=x}$\label{fig:Htrain}}
\end{figure}

In the OVA case, we build $Q_{Y\mid X=x}^{\mathrm{OVA}}$ from a normalization of a set of binary classifiers $\{q_{A_i\mid X=x}\}_{i=0}^{K-1}$. Every $q_{A_i\mid X=x}$ is trained on the set $\{(x_n,a_n)\}_{n=1}^{N}$ where $a_n=\indfunc{y_n=i}$, namely, an indicator for the class $i$. For the hierarchical classification method,  the binary classifier of node $i$ denoted by $q_{S_i\mid X=x}$, is built from the subset of $T$ whose labels $y_i$ belong to the set $S_i$. Then, samples with labels in $S_i^1$ get the binary value one, and samples with labels in $S_i^0$ get the binary value zero. This concept is illustrated in Fig.~\ref{fig:Htrain} where $q_{S_1\mid X=x}$ is trained using the samples with labels 
$0$, $1$, $2$ and sampled labeled $0$, $1$ are trained against samples labeled $2$.

Note that for the training process of the binary classifier at node $i$ we only use samples in the training set $T=\{(x_i,y_i)\}_{n=1}^N$ for which $Y_n\in S_i$. Thus, the samples used for training that classifier are i.i.d. samples from $P_{XY|Y\in S_i}$. Recalling eq.~\eqref{eq:Ltreecond} from Corollary~\ref{cor:treecond}, we see that the contribution of node $i$ to the total regret $R(P_{Y|X},Q_{Y|X}|P_X)$ is determined by $r({p_{S_i\mid X}},{q_{S_i\mid X}}\mid P_{X\mid Y\in S_i})$, which also depends only on the conditional distribution $P_{XY|Y\in S_i}$. Consequently, the expression for the total regret in Corollary~\ref{cor:treecond} is aligned with the training procedure we employ, and the total regret will indeed only depend on how well we can train each of the binary classifiers at each node $i=0,\ldots,K-2$ for minimizing the regret with respect to the distribution $P_{XY|Y\in S_i}$.

\section{{A Leveraged Hierarchical Classifier}\label{sec:leveraging}}
For the hierarchical classification method we introduce another training method in addition to the straightforward ``black-box" method presented above. It is termed \textit{leveraged} training and produces \textit{leveraged} hierarchical classifiers. The basic idea is to
start from a multiclass parametric model $Q_{Y|X;\theta}$, where $\theta\in\Theta$ is the set of parameters determining the conditional probability assignment, and ``project'' it to each node in the hierarchical classification tree from Section~\ref{sec:gentree}, such that now each node consists of a binary classifier whose probability assignment is dictated by the parameters $\theta\in\Theta$. In the original parametric multiclass with the probability assignment $Q_{Y|X;\theta}$, the classifiers at all nodes are dictated by the \emph{same} set of parameters $\theta\in\Theta$. However, in light of Corollary~\ref{cor:treecond}, we see that there is no coupling between the contributions of the different nodes to the total regret. Thus, we can reduce the total regret by reducing each of the individual node regrets, and the latter can be attained by allowing a \emph{different} set of parameters $\theta_i\in\Theta$ to each of the nodes $i\in\{0,\ldots,K-2\}$.

To be more specific, let $Q_{Y|X;\theta}$ be a parametric family of conditional distributions, one for each $\theta\in\Theta$. To fix ideas, think of $Q_{Y|X;\theta}$ as the Softmax probability assignment, as given in~\eqref{eq:softmaxdef} below.
For any specific hierarchical classification tree structure (suitably pruned to have exactly $K-1$ internal nodes) the multiclass probabilities ${P}_{Y\mid X}$, and~$Q_{Y\mid X;\theta}$ induce the set of Bernoulli probabilities $\{p_{S_i\mid X}\}_{i=0}^{K-2}$ and ~$\{q_{S_i\mid X;\theta}\}_{i=0}^{K-2}$, respectively.
To express these probability assignments explicitly, we define
\begin{align}
P_{Y\mid X=x}(S)\dfn \Pr(Y\in S\mid X=x)
\end{align}
and denote its estimate by $Q_{Y\mid X=x;\theta}(S)$. Then, the induced  Bernoulli probabilities are
\begin{align}
p_{S_i\mid X=x}&=\frac{P_{Y\mid X=x}(S_i^1)}{P_{Y\mid X=x}(S_i)}\\
&=\frac{\sum_{j\in S_i^1}{P}_{Y\mid X=x}(j)}{\sum_{j\in S_i}{P}_{Y\mid X=x}(j)} \label{eq:Psibin}
\end{align}
and 
\begin{align}
q_{S_i\mid X=x;\theta}&=\frac{Q_{Y\mid X=x;\theta}(S_i^1)}{Q_{Y\mid X=x;\theta}(S_i)}\\
&=\frac{\sum_{j\in S_i^1}{Q}_{Y\mid X=x;\theta}(j)}{\sum_{j\in S_i}{Q}_{Y\mid X=x;\theta}(j)}\label{eq:qbin}
\end{align}
for $i\in\{0,...,K-2\}$. 
We view the expression for $q_{S_i\mid X=x;\theta}$ above as the ``projection'' of the multiclass parametric family $Q_{Y|X;\theta}$ to the binary classifier at the $i$th node in the tree. This projected binary classifier itself evidently belongs to a parametric family corresponding to $\theta\in \Theta$.
Now, applying Corollary~\ref{cor:treecond} we see that the regret corresponding to $Q_{Y|X;\theta}$ is given by
\begin{align}
&{R}({P}_{Y\mid X}, Q_{Y\mid X;\theta}\mid P_X)=\\
&\sum_{i=0}^{K-2}\Pr(Y\in S_i)r({p_{S_i\mid X}},{q_{S_i\mid X;\theta}}\mid P_{X\mid Y\in S_i}).\label{eq:Ltreecondtheta}
\end{align}
The regret is clearly a function of the chosen parameter $\theta\in \Theta$, and empirical risk minimization (ERM) training chooses the value of $\theta\in\Theta$ that minimizes~\eqref{eq:Ltreecondtheta} where $P_{XY}$ is replaced with the empirical distribution induced by the training set $T=\{(x_i,y_i)\}_{n=1}^N$.

So far, we have only provided an expression for the regret attained by the parametric class $Q_{Y|X;\theta}$ in terms of the hierarchical classification tree from Section~\ref{sec:gentree}. However, once the expression~\eqref{eq:Ltreecondtheta} is established, it is immediately evident that the total regret can be reduced if we allow each binary classifier at the nodes $i=\{0,\ldots,K-2\}$ to use a different set of parameters $\theta_i\in \Theta$, rather than constraining all nodes to use the same $\theta\in \Theta$, as in the multiclass family $Q_{Y|X;\theta}$ we have started with. Our leveraged hierarchical classification method therefore \emph{separately minimizes} each of the binary regret terms in~\eqref{eq:Ltreecondtheta}. In particular, if ERM is used for training, we choose $\theta_i\in\Theta$ as the minimizer of $r({p_{S_i\mid X}},{q_{S_i\mid X;\theta}}\mid P_{X\mid Y\in S_i})$ with respect to the empirical distribution $P_{XY|Y\in S_i}$ induced by the training set. Since the minimization here is on a larger parametric space, the empirical regret can only decrease (or remain the same). The generalization error may increase due to the larger number of parameters, but our experiments below indicate that often the total regret attained by the leveraged classifier is significantly reduced with respect to the original multiclass classifier.

Let us now demonstrate the leveraged hierarchical classification method for the important special case where the baseline multiclass classifier is logistic regression (Softmax).
In this case, $\theta$ is the set of vectors $\{\beta_j\}_{j=0}^{K-1}$ and
\begin{align}
Q_{Y\mid X=x;\theta}(j)&=\frac{\exp(\beta^T_j x)}{\sum_{\ell=0}^{K-1}\exp(\beta^T_\ell x)}\label{eq:softmaxdef}
\end{align}
where we did not use the convention that $\beta_0$ is the all-zeros vector, and assumed the intercept was handled by adding a constant coordinate to $x$. The induced conditional binary classifiers are now
\begin{align}\label{eq:qMOVA}
q_{S_i\mid X=x;\theta}=
\frac{\sum_{j\in S_i^1}\exp(\beta^T_j x)}{\sum_{j\in S_i}\exp(\beta^T_j x)}.
\end{align}

We now allow every binary classifier to have a distinct parameter set $\theta_i$ which comprises the vectors $\{\gamma_{ij}\}_{j\in S_i}$. Namely
\begin{align}
q_{S_i\mid X=x;\theta_i}=
\frac{\sum_{j\in S_i^1}\exp(\gamma^T_{ij} x)}{\sum_{j\in S_i}\exp(\gamma^T_{ij} x)}.
\end{align}
The empirical log-loss function related to the node associated with the set $S_i$ is therefore
\begin{align}\label{eq:LoglosH} 
&\hat{L}(T,\{\gamma_{ij}\}_{j\in S_i})=
\sum_{n:y_n \in S_i}\log\left[\sum_{j\in S_i}e^{\gamma^T_{ij} x_n}\right]-\\
&\sum_{n:y_n \in S_i^{0}}\log\left[\sum_{j\in S_i^{0}}e^{\gamma^T_{ij} x_n}\right]
-\sum_{n:y_n \in S_i^{1}}\log\left[\sum_{j\in S_i^{1}}e^{\gamma^T_{ij} x_n}\right].
\end{align}
This function can be optimized using stochastic gradient descent (SGD) using the derivatives
\begin{align}
&\frac{\partial}{\partial \gamma_{ij}}\hat{L}(T,\{\gamma_{ij}\}_{j\in S_i}) \\
&=\sum_{n:y_n \in S_i}\frac{x_n e^{\gamma_{ij} x_n}}{\sum_{k\in S_i}e^{\gamma_{ik} x_n}}
-
\sum_{n:y_n \in S^0_i}\frac{x_n e^{\gamma_{ij} x_n}}{\sum_{k\in S^{0}_i}e^{\gamma_{ik} x_n}}\label{eq:sgdder1}
\end{align}
for $j\in S_i^0$ and 
\begin{align}
&\frac{\partial}{\partial \gamma_{ij}}\hat{L}(T,\{\gamma_{ij}\}_{j\in S_i}) \\
&=\sum_{n:y_n \in S_i}\frac{x_n e^{\gamma_{ij} x_n}}{\sum_{j\in S_i}e^{\gamma_{ij} x_n}}
-
\sum_{n:y_n \in S^1_i}\frac{x_n e^{\gamma_{ij} x_n}}{\sum_{k\in S^{1}_i}e^{\gamma_{ik} x_n}}\label{eq:sgdder2}
\end{align}
for $j\in S_i^1$.

{It is in place to note that adding a regularization penalty term to the empirical loss (such as the $L_2$ norm of the parameters vectors) can potentially improve the generalization error. However, the concept of regularization is left out of the scope of this paper.}
The experimental results using this method are given in the following section. 


\section{Experimental Results\label{sec:numeric}}
\subsection{Synthetic Gaussian Data}

\begin{table*}[ht]
\centering
{
\begin{tabular}{|c|c||c|c|c|c|c|c|c|c|}
	\hline
	Scenario& N& \multicolumn{2}{ c|}{Softmax}  &  \multicolumn{2}{ c|}{OVA}  &  \multicolumn{2}{ c|}{COVA}  &  \multicolumn{2}{ c|}{LCOVA}  \\
	&& Train & Test & Train & Test & Train & Test & Train  &Test\\
	\hline\hline
	A&$10^5$ &
	$-0.0044$& $0.0049$ &
	$0.0012$&$0.0099$& 
	$0.0123$&$0.0242$ & 
	$-0.0221$&$0.0179$ \\
	\hline
	A&$10^6$ &
	$-0.0004$&$0.0004$  &
	$0.0052$&$0.0060$& 
	$0.0178$&$0.0181$& 
	$-0.0011$&$0.0037$ \\
	\hline
    B&$10^5$ &
	$0.7010$ &$0.7018$  &
	$0.7062$&  $0.7063$& 
	$0.7153$& $0.7189$ & 
	$0.6456$&$0.6786$ \\
	\hline
	B&$10^6$ &
	$0.7012$ &$0.7039$  &
	$0.7063$&  $0.7088$& 
	$0.7172$& $0.7193$ & 
	$0.6519$&$0.6606$ \\
	\hline
\end{tabular}
}
\caption{Experimental results. The entries represent regret values in natural logarithm \label{tab:experiemnts}}
\end{table*}
The first set of experimental results we present were obtained using synthetic data drawn from Gaussian distributions. There are two motivations for the use of synthetic data. The first is the ability to calculate the true sample probabilities and the associated log-loss, which in turn, enables the calculation of the regret terms. The second is the ability to control the number of training samples and therefore to evaluate the sample size vs.~generalization error trade-off. The synthetic data comprise ten equiprobable classes of dimension $100$. In the experiments, the conditional probability density function associated with the $k$th class is multi-dimensional Gaussian. Namely, $[X|Y=i]\sim\mathcal{N}(\mu_i,\Sigma_i)$. We tested two scenarios. In both, the class centers $\{\mu_i\}$ are distinct with all elements independently drawn from a Gaussian distribution. In Scenario A, all the class covariance matrices are $\Sigma_i=\sigma^2{I}$. In Scenario B $\Sigma_i=\sigma^2{I}+\alpha\cdot A_i^TA_i$ where $A_i$ are class independent $d\times d$ matrices whose elements are independently drawn from a Gaussian distribution, and $\alpha$ is a scaling factor set to $0.1$. The class centers and covariances are drawn and held constant while generating the training set and the test set.

The experiments were conducted using a Python environment \cite{MulticlassCode}. The Softmax classifier and the binary logistic regression classifiers were trained using a standard Python implementation \cite{scikit-learn}. 
The COVA classifier is a specific hierarchical classifier that we discussed in Subsection~\ref{subsec:cova}. LCOVA, is its leveraged version explained Section~\ref{sec:training} and trained using SGD with the derivatives given in \eqref{eq:sgdder1} and \eqref{eq:sgdder2}.
The results are summarized in Table~\ref{tab:experiemnts}. $N$ is the number of training samples, and the regrets are calculated w.r.t.~the minimal log-loss, evaluated on the sample using the knowledge of the true distribution. 
For Scenario A, the near zero regrets of Softmax can be explained by the fact that the Softmax (i.e. multiclass logistic regression) posteriors have the exact same form of the ones of the additive Gaussian model with class independent noise covariance matrix (in fact, logistic regression is a more general model than additive Gaussian, see for example the discussion in \cite[Subsection~4.4.5]{hastie2009elements}). As explained in Section~\ref{sec:leveraging}, LCOVA generalizes Softmax in the sense that it has the same node Bernoulli probabilities but without the constraint that all nodes share the same parameters. However, LCOVA has a larger number of parameters than Softmax, which explains the overfitting. Namely, for the same number of training samples it has lower log-loss (and regret) on the training set, but a larger log-loss on the test set.
\newline
For Scenario B, the Softmax parametric family does not include the correct posterior probability assignment $P_{Y|X}$, which explains its increased regret terms. In this case LCOVA, being a richer model that generalizes both COVA and Softmax, provides lower regrets. However, due to its larger number of parameters it does exhibit a larger generalization error than Softmax. As expected the generalization error diminishes with the increase in the number of training samples. Nevertheless, even for $N=10^5$, despite the generalization error, LCOVA still outperforms Softmax.

\subsection{MNIST Database}
In a second set of experiments we used the popular MNIST database
\cite{deng2012mnist}, which comprises gray-scale images of handwritten digits. The corpus contains $60,000$ training samples and $10,000$ test samples. The image size is $28\times 28$ pixels, and we standardly cropped the margins to produce $20\times 20$ pixel images. 

\begin{figure}[h]
	\centering
	\includegraphics{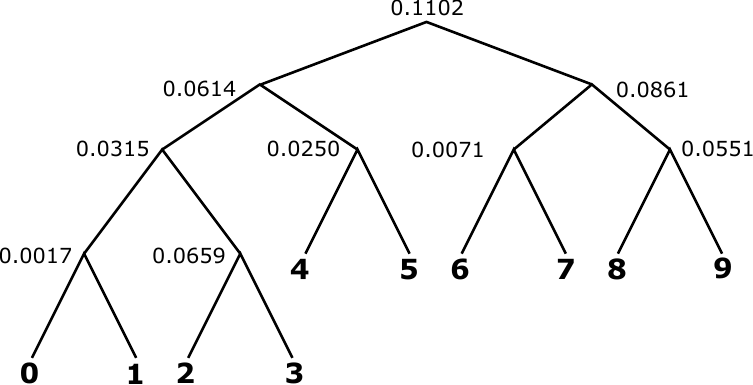}
	\caption{MNIST digits hierarchical decision tree. The numbers on the nodes are the associated test log-loss of the leveraged hierarchical classifier \label{fig:hmnist}}
\end{figure}

The results are presented in Table~\ref{tab:MNIST}. The third row in the table corresponds to a hierarchical classifier with binary logistic regression classifier at the tree nodes (which is equivalent to hierarchical Softmax of
\cite{morin2005hierarchical}) based on the tree of Fig.~\ref{fig:hmnist}. The fourth row is a leveraged version of this hierarchical classifier. We can see that while the hierarchical classifier performs poorly w.r.t.~the Softmax baseline, its leveraged version provides a dramatic improvement in terms of both error rate and log-loss on both the training set and the test set. The numbers written near the nodes of Fig.~\ref{fig:hmnist} are the binary log-losses calculated on the test set. These log-losses sum up to the total multiclass log-loss using the appropriate class weighing.
 
\begin{figure}[h]
	\centering
	\includegraphics{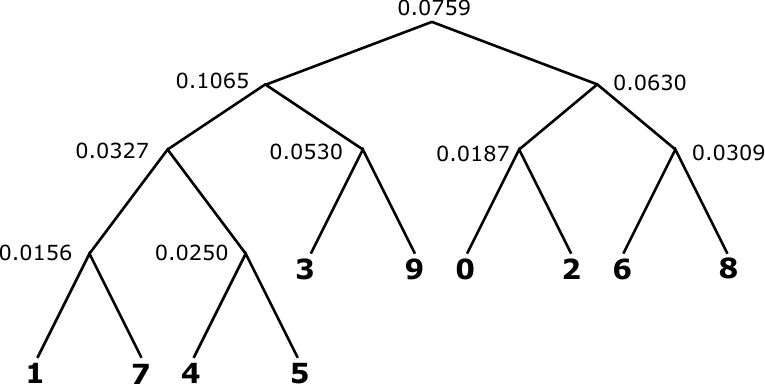}
	\caption{A different hierarchical decision tree for the MNIST digits. The numbers on the nodes are the associated test log-loss of the leveraged hierarchical classifier \label{fig:hmnistPerm}} 
\end{figure}

We note that the tree structure can dramatically change the performance of the hierarchical and leveraged hierarchical classifier. For example, the last row in Table~\ref{tab:MNIST} corresponds to the tree in Fig.~\ref{fig:hmnistPerm}. The motivation behind this tree is to separate digits according to their graphical representation. For example, the root separates between the curved digits $\{0,2,6,8\}$ and non-curved digits $\{1,7,4,5,3,9\}$. As can be seen from the table, the performance of this tree is better than the baseline performance, and may potentially be improved further by using different tree structures.

\begin{table}[ht]
\centering
\begin{tabular}{|c||c|c|c|c|}
	\hline
	Method & 
	\multicolumn{2}{ c|}{Error }  &  \multicolumn{2}{ c|}{log-loss}\\    
	 & Train & Test & Train & Test \\
	\hline \hline
	Softmax &$7.0\%$ &$7.9\%$ &$0.253$ &$0.290$ \\
	\hline
	OVA& $8.0\%$& $8.4\%$& $0.347$ & $0.368$ 
	\\
	\hline
	Hierarchical&$16.8\%$ &$17.3\%$ & $0.543$& $0.571$\\
	\hline
	L-Hierarchical& $4.5\%$&$6.0\%$  &$0.154$ & $0.225$ \\
	\hline
	L-Hierarchical& $4.1\%$&$5.6\%$  &$0.141$ & $0.207$ \\
	permuted& & & & \\
	\hline
\end{tabular}
\caption{MNIST results}
\label{tab:MNIST}
\end{table}

\subsection{Discussion}
The results presented in the previous subsection show the potential of leveraged hierarchical classification to improve upon Softmax. The following aspects can be inferred from the simulation results and are left for further research
\begin{enumerate}
\item \underline{{The} hierarchical tree structure:} As mentioned in Section~\ref{sec:training}, for any tree structure, a multiclass classifier can be factorized into binary elements using \eqref{eq:Psibin}, \eqref{eq:qbin}. Then, the binary log-losses will sum up to the multiclass regret, given the appropriate weighing. However, independent training of the binary classifier can yield different log-losses and total performance as seen in the MNIST experiments. A judicious tree architecture can either be based on some knowledge of the nature of the data (a suggested for natural language processing in \cite{morin2005hierarchical}), or can be set on-the-fly using a different algorithm, which is out of the scope of this contribution.
\item \underline{{The} number of parameters:} A standard Softmax classifier is characterized by a set of $K$ vectors of size $d$: $\{\beta_i\}_{i=0}^{K-1}$. In a hierarchical Softmax based tree, the number of vectors in each node is equal to the number of members in its class subset $S_i$. Therefore, the total number of vectors in a COVA tree is $\approx K^2/2$ and for a balanced hierarchical tree with depth $\approx \log K $ it is $\approx K\log K$. While a COVA tree might better suite specific data structure, it is clear that a balanced tree is preferable in terms of complexity (both training and probability assignment) and can potentially yield a smaller generalization gap due to its smaller number of parameters. 

That being said, in the regime where deep neural nets (DNNs) are used, one often has a number of training samples far exceeding the number of classes. Consequently, replacing the last layer in a DNN with a hierarchical classifier (that has a larger number of parameters, but more flexibility) might still yield improved results.
\newline
Even if $P_{XY}$ is such that there exist weights for which Softmax provides small log-loss, during training (before the weights are well tuned) the log-loss for the $Q_{Y|X}$ output by the DNN classifier is large. Since the leveraged hierarchical classifier can offer significant gains in this regime, we suspect that using it instead of Softmax may lead to a speed-up in the training time.    
\item \underline{Regularization:} The results in Table~\ref{tab:MNIST} show a consistent gap between train and test performance for all methods of both error and log-loss. All methods in this table are non-regularized and it makes sense that standard regularization (using for example L1 or L2 penalty) might reduce the generalization gap. We regard the main contribution in this paper as theoretical, providing qualitative tools to analyze the log-loss of various classifier architecture. The experimental part is deliberately concise and is designated to demonstrate the theoretic tools, but not to optimize the result nor to compete with the vast body of work in the area. For this reason, we did not apply regularization in the experiments. 
\end{enumerate}

\section{Concluding Remarks\label{eq:conclusion}}
We have studied the problem of soft classification under log-loss and have focused on constructions of multiclass classifiers from binary classifiers. For the popular one vs. all method, we have shown that the total regret is upper bounded by the sum of regrets corresponding to the underlying binary classifiers. We have then considered the hierarchical classification method, and derived an exact expression for the total regret of this method in terms of the regrets corresponding to the binary classifiers used at each node of the hierarchical classification tree.

Building on this expression, we have noticed that the performance of every multiclass classifier is solely dictated by that of the binary classifiers at each node of that tree. This observation suggests that improved performance may be obtained by optimizing the ``projected'' binary classifiers separately. We have provided some numerical evidence showing that this is often indeed the case.

Our analysis heavily relied on properties special to the logarithmic loss. However, using Pinsker's inequality we also applied our results for obtaining bounds on the zero-one loss. {Furthermore, we note that  cross-entropy is commonly used as the loss function for the training process even when performance is eventually assessed using different losses. In these cases, the clean and elegant nature of the regret expressions we obtained under logarithmic loss shed much insight on the terms dominating the training procedure.}


\section{Acknowledgment}
The authors would like to thank Meir Feder and Yury Polyanskiy for many fruitful discussions.



\bibliographystyle{IEEEtran}

\begin{thebibliography}{10}
\providecommand{\url}[1]{#1}
\csname url@samestyle\endcsname
\providecommand{\newblock}{\relax}
\providecommand{\bibinfo}[2]{#2}
\providecommand{\BIBentrySTDinterwordspacing}{\spaceskip=0pt\relax}
\providecommand{\BIBentryALTinterwordstretchfactor}{4}
\providecommand{\BIBentryALTinterwordspacing}{\spaceskip=\fontdimen2\font plus
\BIBentryALTinterwordstretchfactor\fontdimen3\font minus
  \fontdimen4\font\relax}
\providecommand{\BIBforeignlanguage}[2]{{%
\expandafter\ifx\csname l@#1\endcsname\relax
\typeout{** WARNING: IEEEtran.bst: No hyphenation pattern has been}%
\typeout{** loaded for the language `#1'. Using the pattern for}%
\typeout{** the default language instead.}%
\else
\language=\csname l@#1\endcsname
\fi
#2}}
\providecommand{\BIBdecl}{\relax}
\BIBdecl

\bibitem{daniely2012multiclass}
A.~Daniely, S.~Sabato, and S.~S. Shwartz, ``Multiclass learning approaches: A
  theoretical comparison with implications,'' \emph{arXiv preprint
  arXiv:1205.6432}, 2012.

\bibitem{rifkin2004defense}
R.~Rifkin and A.~Klautau, ``In defense of one-vs-all classification,''
  \emph{The Journal of Machine Learning Research}, vol.~5, pp. 101--141, 2004.

\bibitem{hastie1998classification}
T.~Hastie, R.~Tibshirani \emph{et~al.}, ``Classification by pairwise
  coupling,'' \emph{Annals of statistics}, vol.~26, no.~2, pp. 451--471, 1998.

\bibitem{dietterich1994solving}
T.~G. Dietterich and G.~Bakiri, ``Solving multiclass learning problems via
  error-correcting output codes,'' \emph{Journal of artificial intelligence
  research}, vol.~2, pp. 263--286, 1994.

\bibitem{allwein2000reducing}
E.~L. Allwein, R.~E. Schapire, and Y.~Singer, ``Reducing multiclass to binary:
  A unifying approach for margin classifiers,'' \emph{Journal of machine
  learning research}, vol.~1, no. Dec, pp. 113--141, 2000.

\bibitem{kumar2002hierarchical}
S.~Kumar, J.~Ghosh, and M.~M. Crawford, ``Hierarchical fusion of multiple
  classifiers for hyperspectral data analysis,'' \emph{Pattern Analysis \&
  Applications}, vol.~5, no.~2, pp. 210--220, 2002.

\bibitem{chen2004integrating}
Y.~Chen, M.~M. Crawford, and J.~Ghosh, ``Integrating support vector machines in
  a hierarchical output space decomposition framework,'' in \emph{IGARSS 2004.
  2004 IEEE International Geoscience and Remote Sensing Symposium},
  vol.~2.\hskip 1em plus 0.5em minus 0.4em\relax IEEE, 2004, pp. 949--952.

\bibitem{vural2004hierarchical}
V.~Vural and J.~G. Dy, ``A hierarchical method for multi-class support vector
  machines,'' in \emph{Proceedings of the twenty-first international conference
  on Machine learning}, 2004, p. 105.

\bibitem{lorena2008review}
A.~C. Lorena, A.~C. De~Carvalho, and J.~M. Gama, ``A review on the combination
  of binary classifiers in multiclass problems,'' \emph{Artificial Intelligence
  Review}, vol.~30, no. 1-4, p.~19, 2008.

\bibitem{delmoral2021pitfalls}
P.~del Moral, S.~Nowaczyk, A.~Sant'Anna, and S.~Pashami, ``Pitfalls of
  assessing extracted hierarchies for multi-class classification,'' \emph{arXiv
  preprint arXiv:2101.11095}, 2021.

\bibitem{morin2005hierarchical}
F.~Morin and Y.~Bengio, ``Hierarchical probabilistic neural network language
  model,'' in \emph{International workshop on artificial intelligence and
  statistics}.\hskip 1em plus 0.5em minus 0.4em\relax PMLR, 2005, pp. 246--252.

\bibitem{silla2011survey}
C.~N. Silla and A.~A. Freitas, ``A survey of hierarchical classification across
  different application domains,'' \emph{Data Mining and Knowledge Discovery},
  vol.~22, no.~1, pp. 31--72, 2011.

\bibitem{merhav1998universal}
N.~Merhav and M.~Feder, ``Universal prediction,'' \emph{IEEE Transactions on
  Information Theory}, vol.~44, no.~6, pp. 2124--2147, 1998.

\bibitem{cesabianchilugosi}
N.~Cesa-Bianchi and G.~Lugosi, \emph{Prediction, learning, and games}.\hskip
  1em plus 0.5em minus 0.4em\relax Cambridge University Press, 2006.

\bibitem{cw14}
T.~A. Courtade and T.~Weissman, ``Multiterminal source coding under logarithmic
  loss,'' \emph{IEEE Transactions on Information Theory}, vol.~60, no.~1, pp.
  740--761, 2014.

\bibitem{jcvw15}
J.~Jiao, T.~A. Courtade, K.~Venkat, and T.~Weissman, ``Justification of
  logarithmic loss via the benefit of side information,'' \emph{IEEE
  Transactions on Information Theory}, vol.~61, no.~10, pp. 5357--5365, October
  2015.

\bibitem{fogel2018universal}
Y.~Fogel and M.~Feder, ``Universal batch learning with log-loss,'' in
  \emph{2018 IEEE International Symposium on Information Theory (ISIT)}.\hskip
  1em plus 0.5em minus 0.4em\relax IEEE, 2018, pp. 21--25.

\bibitem{xr20}
A.~Xu and M.~Raginsky, ``Minimum excess risk in bayesian learning,''
  \emph{arXiv preprint arXiv:2012.14868}, 2020.

\bibitem{deng2012mnist}
L.~Deng, ``The mnist database of handwritten digit images for machine learning
  research,'' \emph{IEEE Signal Processing Magazine}, vol.~29, no.~6, pp.
  141--142, 2012.

\bibitem{CoverThomas}
T.~M. Cover and J.~Thomas, \emph{{Elements of Information Theory}},
  2nd~ed.\hskip 1em plus 0.5em minus 0.4em\relax New York: Wiley, 2006.

\bibitem{yeung1991local}
R.~W. Yeung, ``Local redundancy and progressive bounds on the redundancy of a
  huffman code,'' \emph{IEEE transactions on information theory}, vol.~37,
  no.~3, pp. 687--691, 1991.

\bibitem{MulticlassISIT}
A.~Ben-Yishai and O.~Ordentlich, ``Constructing multiclass classifiers using
  binary classifiers under log-loss,'' in \emph{2021 IEEE International
  Symposium on Information Theory (ISIT)}, 2021, pp. 2435--2440.

\bibitem{bnop18}
A.~Bhatt, B.~Nazer, O.~Ordentlich, and Y.~Polyanskiy, ``Information-distilling
  quantizers,'' \emph{IEEE Trans. Inf. Theory}, accepted, 2021.

\bibitem{ot2020}
O.~Ordentlich and I.~Tal, ``An upgrading algorithm with optimal power law,''
  \emph{arXiv preprint arXiv:2004.00869}, 2020.

\bibitem{MulticlassCode}
\BIBentryALTinterwordspacing
{A.~Ben{-}Yishai and O.~Ordentlich}. {Multiclass using binary classifiers
  Python code}. [Online]. Available:
  \url{{https://github.com/assafbster/MulitClass-clssifiers}}
\BIBentrySTDinterwordspacing

\bibitem{scikit-learn}
F.~Pedregosa, G.~Varoquaux, A.~Gramfort, V.~Michel, B.~Thirion, O.~Grisel,
  M.~Blondel, P.~Prettenhofer, R.~Weiss, V.~Dubourg, J.~Vanderplas, A.~Passos,
  D.~Cournapeau, M.~Brucher, M.~Perrot, and E.~Duchesnay, ``Scikit-learn:
  Machine learning in {P}ython,'' \emph{Journal of Machine Learning Research},
  vol.~12, pp. 2825--2830, 2011, \url{https://scikit-learn.org/stable/}.

\bibitem{hastie2009elements}
T.~Hastie, R.~Tibshirani, and J.~Friedman, \emph{The elements of statistical
  learning: data mining, inference, and prediction}.\hskip 1em plus 0.5em minus
  0.4em\relax Springer Science \& Business Media, 2009.

\end{thebibliography}

\end{document}